\documentclass[numbered]{trbunofficial}
\usepackage{graphicx}
\usepackage{bchart}
\usepackage{subcaption}
\usepackage{caption}
\usepackage{pgfplots}
\usepackage{pgfplots}

\usepackage{amsmath}
\usepackage[hidelinks]{hyperref}

\AuthorHeaders{Ajit, Mouli, Knickerbocker, and Wood}

\title{Machine learning framework for end-to-end implementation of Incident duration prediction}

\author{%
  \textbf{Smrithi Ajit\\
   smrithiajit@gmail.com}\\
  \hfill\break
  \textbf{Varsha R Mouli, Research Engineer, Institute of Transportation\\
  varsha@iastate.edu}\\
  \hfill\break%
  \textbf{Skylar Knickerbocker, Research Engineer, Institute of Transportation\\
  sknick@iastate.edu}\\
  \hfill\break%
  \textbf{Jonathan S. Wood, Ph.D, Assistant Professor, Dept.of Civil,Construction and Environmental Engineering, Iowa State University\\
  jwood2@iastate.edu}
}

\begin{document}

\maketitle

\section{Abstract}
Traffic congestion caused by non-recurring incidents such as vehicle crashes and debris is a key issue for Traffic Management Centers (TMCs). Clearing incidents in a timely manner is essential for improving safety and reducing delays and emissions for the traveling public. However, TMCs and other responders face a challenge in predicting the duration of incidents (until the roadway is clear), making decisions of what resources to deploy difficult. To address this problem, this research developed an analytical framework and end-to-end machine-learning solution for predicting incident duration based on information available as soon as an incident report is received. Quality predictions of incident duration can help TMCs and other responders take a proactive approach in deploying responder services such as tow trucks, maintenance crews or activating alternative routes. The predictions use a combination of classification and regression machine learning modules. The performance of the developed solution has been evaluated based on the Mean Absolute Error (MAE), or deviation from the actual incident duration as well as Area Under the Curve (AUC) and Mean Absolute Percentage Error (MAPE). The results showed that the framework significantly improved incident duration prediction compared to methods from previous research.

\hfill\break%
\noindent\textit{Keywords}: non-recurring, crash, predicts, incident duration, machine learning, MAE 
\newpage

\section{Introduction and Background}

According to the Traffic Incident Management Handbook, an incident is defined as a non-recurring event resulting in either a reduction in roadway capacity or an abnormal increase in demand \cite{58}. These incidents include, but are not limited to, vehicle crashes, disabled vehicles, debris, and spilled cargo. Incidents not only result in traveler delay but also increase the likelihood of secondary crashes and other secondary effects due to increased opportunities for secondary events to occur \cite{59}. Secondary events can lead to increased demand for police, fire, and emergency services, reduced air quality, and other environmental impacts.

\begin{figure}[h]
  
\includegraphics[width=\textwidth,height=8cm]{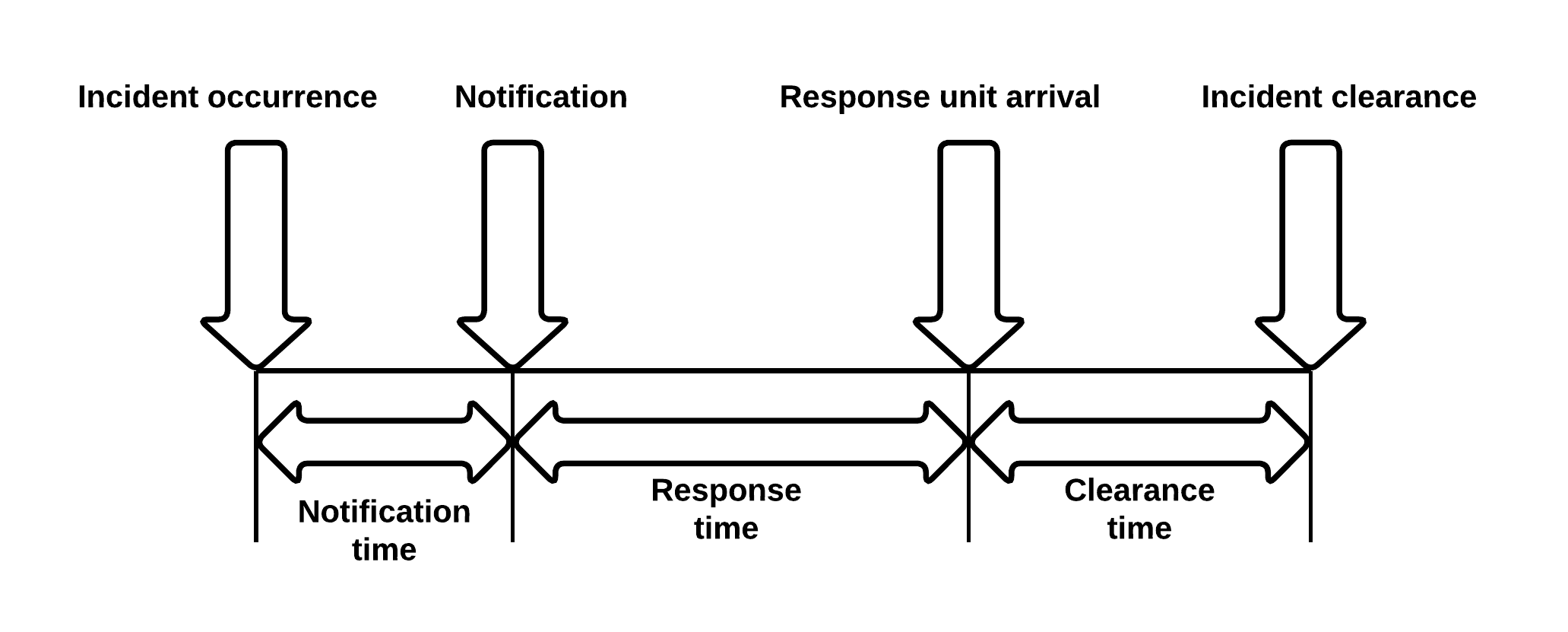}

  \caption{Illustration of traffic incident duration }
    \label{fig:incidentduration}
\end{figure}

Total incident duration is comprised of incident notification time, response time, and clearance time, as illustrated in Figure~\ref{fig:incidentduration}. As shown, the total incident duration is the total time from the start of the incident until the reported time of clearance for the event \cite{2} and includes incident notification, response, and clearance times. The incident notification time is from the start of the incident until the time it is reported. The response time is from the report time until the response unit's arrival. The clearance time is the time taken to clear the incident after emergency responders have arrived on the scene.

While the incident duration time is not controlled by the responding agencies, dispatching the correct personnel and equipment can reduce the total incident duration by minimizing the sum total of response and clearance times. This requires planning, preparedness, and coordination between responders. Information on what resources should be dispatched can be improved using predictive models of the total incident duration. For instance, having accurate predictions of the total incident duration can assist Traffic Management Center (TMC) operators in selecting the appropriate actions from potential options such as the following: (1) diverting traffic to an alternate route, (2) providing a warning of a potential delay to travelers planning to take a congested route and (3) ensuring helper services, such as safety service patrol or maintenance crews, arrive at the incident spot on time. For example, if an incident will be cleared within a half hour, it may not be reasonable to detour traffic on a route that increases the travel time over that amount of time. Helper services may also not be requested if the incident is cleared before the time it takes the service patrol to arrive at the incident location. 

As indicated above, incident duration prediction is critically important to the TMC for timely mitigation of traffic congestion, to not only forewarn people of the crashes on a particular route in advance but reduce the likelihood of secondary crashes. Past studies have focused on understanding and identifying factors related to incident type, roadway data, time of the day, weather conditions, speed, traffic volume, blocked lane, location, environment, weather, road characteristics, temporal and spatial factors, and so on, either through associative mining or through prediction based descriptive models that are either statistical or machine learning based models. 

Early work in the area of incident duration prediction used linear regression models and related statistical tests such as Analysis of Variance (ANOVA). Many of these models were limited by the number of data points \cite{35garib},\cite{37peeta} \cite{42yuxia}. Coefficient of determination, Root Mean Squared Error (RMSE), Mean Absolute Error (MAE), and Mean Absolute Percentage Error (MAPE) were the performance metrics commonly used to report the efficiency of the ML models. The lowest MAPE that was reported with these models include that by \citep{43weng}, who reported 34.1\% for a total of 2,512 accidents using a cluster-based log-normal distribution model and a MAPE of 37\% reported by Khattak et al. for a total of 59,804 accidents. 

Other methods that have been used in the published literature include fuzzy logic, Artificial Neural Networks (ANN), Bayesian methods, Survival or Hazard models, tree-based machine learning methods, and text mining. Among fuzzy logic-based approaches, the lowest MAPE was 36\% with an average error of 0.3 minutes reported by the authors in \cite{44fuzzy} and  \cite{45fuzzykim}, respectively. However, these studies were also limited by fewer data points. Methods using ANN have also been attempted on small and large datasets, with Pereira et.al\cite{47pereira} achieving a median error of 9.9 minutes for a dataset of size 10,139 and Lopes et al. \cite{46lopes} achieving within 10 to 20 minutes error for a dataset of 10,762 incidents.

Bayesian models are useful in applying prior knowledge of associations between variables in developing predictive models. Several authors have developed the incident duration prediction problem into a classification problem and have used Bayesian approaches for predictions achieving accuracy ranging between 74\% to 80\%. Authors of \citep{14} used the Bayesian ANN model for a large dataset achieving a low MAPE of 0.18 to 0.29. Other Bayesian approaches reported include Bayesian propagation neural network, and Bayesian Support Vector Regression \cite{9}.
 
Survival models or Hazard based models are used to predict the length of the duration between the occurrence of the event, and its clearance while also being capable of predicting when the incident duration will end given it has proceeded for time t. This ability of the models helps prediction of incident duration at different stages of the incident as presented in \citep{li2015traffic} using an AFT hazard-based model. Fully parametric Hazard based models using log-normal and log-logistic distributions were used in \citep{chung2010development} and reported a MAPE value of 47\%. A logistic Accelerated Failure Time model (AFT) has been presented in \citep{hu2011incident} that reported a MAPE of 43.7\%, and another Weibull distribution parametric model using gamma heterogeneity was used in \citep{kang2011applying} to account for unobserved heterogeneity. In real-time, as time progresses, the traffic conditions change, and the new set of factors influence the incident duration, due to which time sequential models are useful. Incorporating additional features as the incident progresses can be very helpful in traffic decision-making. Inverse Gaussian frailty AFT model, Multilevel mixed-effect AFT model, and FMAM models were used to predict the incident duration with sequential TIM information in \citep{li2020sequential}.
Multiple studies also identified that a single method could not suit all the incident duration ranges and came up with hybrid approaches that combine two or more methods. In the study \cite{48lin}, Hazard Based Duration Models (HBDMs) were used as leaves of the M5P Tree, and in the work of  \citep{49lin}, fuzzy entropy was used to select the features, and ANN was used to predict the duration based on these features. A competing mixture model was presented in \cite{24} to analyze the influence of the clearance method and other covariates on traffic incidents. 
 
Tree-based models have also received significant attention. Classification Tree methods have reported MAPE values ranging between 42.7\% to 65\% \cite{50crt,51crt,52crt}. A more recent focus has been on ensemble techniques like RF, Gradient Boost method, Extreme Gradient Boost Method, and Ada Boost owing to the superiority of performance with datasets with a large number of categorical variables\cite{53xg,54rf,55adaptive}. In \cite{54rf}, RF models developed for long and short-duration datasets reported an MAE of 36.652 minutes and 14.97 minutes, respectively. 

Besides the approaches discussed above, text-based analysis has also been used. In \cite{1}, the authors illustrated an improvement of more than 35\% over non-textual models, and a knowledge-based method was used in \cite{5} to estimate incident clearance duration on Maryland I-95. In \cite{7}, a spatio-temporal feature learning model called TITAN was proposed that considers hidden spatio-temporal associations by considering connectivity between road segments in addition to identifying high-level features. In \cite{8}, a copula-based tri-variate framework was used to generate a stochastic dependence relationship between the various variables capable of predicting incident duration. 

In the context of ML models, since no prior assumptions are made about the distribution of the data, the ML models will be able to identify underlying relationships only if data challenges like skewness and the heterogenous nature of the data are adjusted for prior to training the ML models. The heterogenous nature of the data is handled through the use of feature selection methods that help restrict the variables to a few most significant ones.
In order to group similar data points, a general approach is to use supervised or unsupervised clustering methods. Supervised clustering of data points followed by application of regression models can be seen in \cite{6} while an unsupervised approach can be seen in \cite{12}. A supervised approach that makes use of blending of results from multiple ML models for the prediction of incident duration has not been applied in the published literature. 

\section{Objectives}
The overall objective of this study is to develop a total incident duration prediction framework to support TMC operations by utilizing machine learning in a two-step process. The first step uses supervised classification to predict which of three time ranges the total incident duration falls within. The ranges include within a half hour (class 1), between half an hour and 2 hours (class 2), and more than 2 hours (class 3) based on \cite{58}. This can be used to assist the operator in identifying incidents that need immediate attention and what type of resources may be required. This step also serves to improve the prediction in the regression module through a prior grouping of similar incidents. The second step utilizes a regression model to produce a more accurate prediction of the incident duration in minutes. Three years of historical incident data are used to train, test, validate, and finalize the machine-learning components used in the framework. Two different feature sets have been considered as part of the analysis. The first one is utilized in the initial prediction and involves basic time and location information while the second feature set is used in the second step as more details are available about the event which includes additional features such as road conditions and environmental factors.  To show that the analysis framework works well even with minimal features, each model framework has been evaluated using both feature sets. An ensemble technique called blending has been utilized to improve the performance of regular ML models used in the framework. The performance of the developed supervised ML framework has also been validated against an alternative framework, an unsupervised ML framework, and a basic regression model that allows censorship, known as a Tobit model.

A few applications of the developed framework have been provided in the Discussion section. Additionally, to understand if the classification step improves predictions, the Mean Absolute Error (MAE) results from the developed framework have been compared with MAE results obtained from a framework that predicts without classifying the data into ranges.

\section{Data Description and Pre-processing}\label{sec:data}

The Advanced Traffic Management System (ATMS) events data is the primary source of information used in this paper. It is maintained by the Iowa Department of Transportation Traffic Management Center (TMC) and contains detailed records of traffic incidents that occurred in the state. The data is collected by TMC operators who are actively managing incidents on the roads and includes information such as the number of lanes closed, start and end times, the presence of emergency responders, and the severity of the incident. This information is useful for creating after-action reports, which allow the Iowa DOT to review and analyze their response to incidents in order to identify areas for improvement. The ATMS data are made available on a daily basis and includes all incidents that have been closed from the previous day \cite{16}. The data used in this study was collected from 2017 to 2019.

The Advanced Traffic Management System (ATMS) data used in this study includes both recurring and non-recurring events, such as work zones, collisions, debris, stopped vehicles, and special events. However, the study focuses specifically on collisions involving 1, 2, or 3 vehicles and debris, as these types of incidents can have a significant impact on traffic flow and are within the control of TMC operators. The study also limits the analysis to incidents that occurred on rural and municipal interstates within the state of Iowa and excludes days with severe weather events as these can significantly impact incident duration. To identify severe weather days, the study used the National Oceanic and Atmospheric Administration's (NOAA) storm database and excluded incidents that occurred in affected counties. After removing incomplete records, the study included a total of 6,832 incidents that occurred over a three-year period.

The data used in the study was imbalanced, with more incidents in the short and medium-duration classes compared to the long-duration class. To address this issue, the study used a method called Synthetic Minority Oversampling Technique (SMOTE), which is a way to balance the data by duplicating samples from the minority class (in this case, the long-duration class) without adding any new information. This helps to ensure that the model is not biased towards the majority class. To address the correlation between variables in the data, the study removed features that had a high correlation value (greater than 0.4) and used the pycaret tool to further address this issue. This helps to ensure that the model is not influenced by correlated features, which can impact the accuracy of the model.

For the analysis, the dependent variable will be the duration of each incident, calculated as the time between when the incident started and when the roadway was cleared. The dataset includes several variables, which are listed in Table \ref{tab:class_noresp_1}, along with their data types and any transformations that were applied. Columns with text data, such as information about responders, were converted to categorical variables. Binary variables were used to encode information about the presence of specific responders (e.g., police, firefighters, etc.) at the incident, with a value of 1 indicating presence and 0 indicating absence. This responder information is described in more detail in Table \ref{tab:respinformation}, which also includes statistics about the duration of incidents for each type of responder.

Categorical variables that were not ordered were one-hot encoded. Missing values for numerical variables were imputed with the mean value while missing values for categorical variables were imputed with a constant based on the mode. For the variables vehicles, trucks, injuries, fatalities, and AADT, the variables were converted to categorical variables as described in Table \ref{tab:class_noresp_1}. The statistics for the Average Annual Daily Traffic (AADT) information in the dataset are provided in Table \ref{tab:aadt}. It can be seen from these statistics that the majority of incidents have a duration in short to medium range.

The basic statistics related to the dataset have been provided in Table \ref{statistics}. These descriptive statistics show the wide variance in roadway clearance times with values ranging from 1 to 1,358 minutes, along with a positive skew of 5.637 minutes. This is addressed by removing the skew in the data using a box cox transformation of the data, which reduces the skew of the dataset to -0.035. 

\begin{table}[t]
  \small
\centering
 \caption{Explanation of variables in the dataset, their datatypes and transformations}
\scalebox{0.8}{
\begin{tabular}{|p{3.5cm}|p{4cm}|p{1cm}|p{1.3cm}|p{7cm}|}
 \hline
 Feature Group & Variable name & Derived & Datatype & Categories\\
\hline
Lane Information & Number of lanes(lanes)& N & Integer & NA\\
& OnlyShouldersClosed & Y & Binary & 1-responded,
0-did not respond\\
\hline
Location information & Direction & N & Binary & 1: North,
2: South,
3: West,
4: East,
\\
 &County region & Y & Factor & North East,
North West,
Central,
South East,
South West\\

& City number & N & Factor & Numerical binning\\  
\hline
Incident& Event Type & Y & Factor & Crash(1vehicle, 2 vehicle and 3 vehicle crash),
Debris\\
\hline
Vehicles involved & Vehicles & N & Factor & 0,1,2 and more than 3 \\
& Trucks & N & Factor & 0,1,2 and more than 3 \\
\hline
Severity & Injuries & N & Factor & 1-injured, 0-not injured \\
& Fatalities & N & Factor & 1-1 or more fatality, 0- no fatality \\
\hline
Responder Information & Responder Type: Police, TOW, DOT, DPS & Y & Factor & 1-responded,
0-did not respond \\
\hline

Detection&Detection method & N & Factor & Police, Highway Helper, Automated, 
DOT, Cameras, Others \\
\hline
Temporal information & Time Of Day(TOD) & Y & Factor & Morning: 7am to 9am,
Early Afternoon:10am to 12pm,
Afternoon: 1pm to 3pm,
Evening rush:4 pm to 6pm,
Evening: 7pm to 9pm,
Night: 10pm to 6 am\\
& Day Of Week(DOW) & Y & Factor & Integer encoding
0 to 6 represent Monday to Sunday\\
& Season & Y & Integer & 1=Winter (December to February),
2=Spring (March to May),
3=Summer (June to August),
4=Autumn (September to November)
\\
& Year & Y & Integer & 2017, 2018, 2019 \\
\hline
Traffic information & Hourly traffic volume & Y & Integer & NA\\
& AADT & Y & Factor & 1:<8,000, 2:8,000-12,000, 3:12,000-24,000, 4:24,000-48,000, 5:>48,000\\

\hline

Road characteristics & Surface width & N & Float & NA\\
& Surface type & Y & Factor & Grade and drained earth, gravel or stone, bituminous over gravel or stone, etc \\
& Terrain & Y & Factor & flat, rolly, hilly\\
\hline
\end{tabular}}

\label{tab:class_noresp_1}
\end{table}

\noindent\begin{table}[t]

\begin{minipage}[t]{.5\textwidth}
\caption{Responder Information} 
\scalebox{0.6}{
\begin{tabular}{|p{2cm}|p{1.8cm}|p{4cm}|p{4cm}|}
 \hline
 Responder  &	\% of total &	Incident duration
 &Incident duration  \\
 (Presence) &incidents& Mean (min) & Median (min)\\
 \hline
DOT-No&	91\%&	39.3&	26
\\
\hline
DOT-Yes&	9\%&	88.7&	61
\\
\hline
EMS-No&	88\%&	38.9&	23
\\
\hline
EMS-Yes&	12\%&	76.2&	57
\\ 
\hline
HH-No&	51\%&	40.0&	22\\
\hline
HH-Yes&	49\%&	47.7&	33\\
\hline
\hline
Police-No&	67\%	&33.6&	17\\
\hline
Police-Yes&	33\%&	65.5&	47\\
\hline
TOW-No&	77\%&	76.3&	19\\
\hline
TOW-Yes&	23\%&	23.7&	51
\\
\hline

\end{tabular}}

\label{tab:respinformation}
\vspace*{0.3 cm}
 \caption{Feature Set Description }
\scalebox{0.6}{\begin{tabular}{|p{4cm} |p{8.7cm}|}
 \hline
 Feature group & Description\\
 \hline
Basic Features (FS1) & Basic temporal and spatial information,vehicles,trucks,injuries 
\\
\hline
Full Features(FS2) &AADT, hourly volume, road and environment factors, Responder information \\
\hline
\end{tabular}}

\label{tab:featuresetinfo} 
\end{minipage}
\begin{minipage}[t]{.5\textwidth}
\caption{Statistical Information}
\scalebox{0.7}{
\begin{tabular}{|p{6.5cm} |p{6cm}|}
 \hline
 Metric & Value(minutes)\\
 \hline
Standard deviation&53.7
\\
Mean & 45.2 \\
Median & 31 \\
Minimum & 1 \\
25th percentile & 10\\
75th percentile & 59\\ 
Maximum & 542 \\
\hline
\end{tabular}}
\label{statistics}
\vspace*{0.85 cm}
\caption{AADT versus Incident Duration} 
\scalebox{0.7}{
\begin{tabular}{|p{2.8cm}|p{3cm}|p{3cm}|p{3cm}|}
 \hline
 AADT & Avg. Total & Median Total & No. of Records \\
 \hline
16,000&	47.8&	28&	153
\\
\hline
27,700&	28.8&	4&	131
\\
\hline
45,300&	54.6&	43&	151
\\
\hline
57,000&	28.4&	14&	266
\\ \hline
100,500&	30.4	&20	&194
\\
\hline
\end{tabular}}
\label{tab:aadt}

\end{minipage}

\end{table}

\begin{figure}[!h]
\centering
  \caption{Overall workflow of developed framework }
\includegraphics[width=0.65\textwidth]{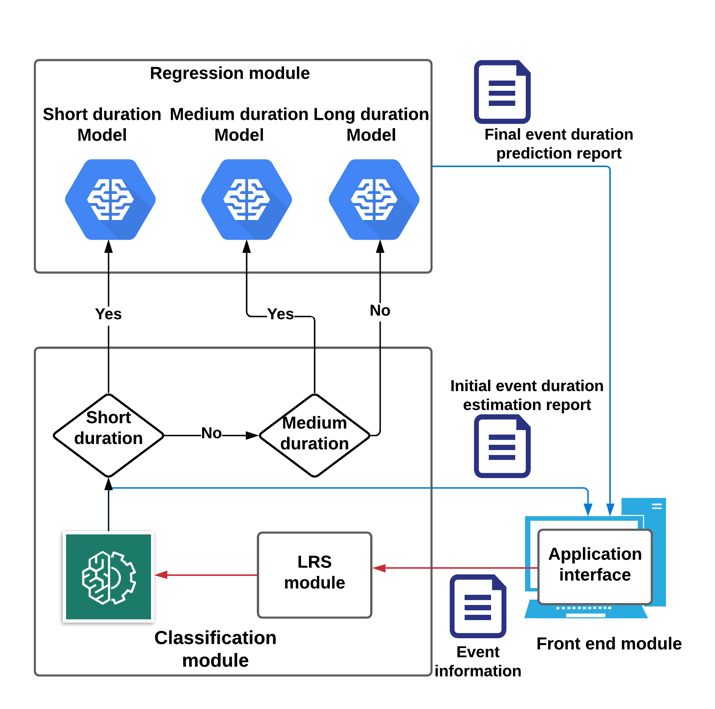}   
  \label{fig:overallworkflow}
\end{figure}

\section{Methodology}
The framework for estimating event duration shown in Figure \ref{fig:overallworkflow} can be separated into two core components, including the classification and regression modules. The goal of the first part of the framework in the classification module is to classify an event as being of short, medium, or long duration based on its total incident duration. Events with a duration of less than 30 minutes are classified as short duration, those between 30 minutes and 2 hours are classified as medium duration, and those lasting more than 2 hours are classified as long duration. 

The initial classification is based on only limited information available about an event, including its location, date, and time of day, since not all details about the event are available when the incident occurs. This minimal information is used to assign the event to one of the three duration categories.

After the initial classification, the data is then processed through a linear referencing system (LRS) module, which removes the spatial dependencies (latitude and longitude coordinates) of the data by adding two additional features: a Route ID and a Measure. The Route ID represents the route on which the event occurred, while the Measure provides the linear distance of the event's location from a reference point. These parameters are used to extract additional features, such as Average Annual Daily Traffic (AADT), terrain, surface width, and surface type, from the Roadway Asset Management System (RAMS). 
\begin{figure}[!h]
  \centering
 \includegraphics[width=0.5\textwidth]{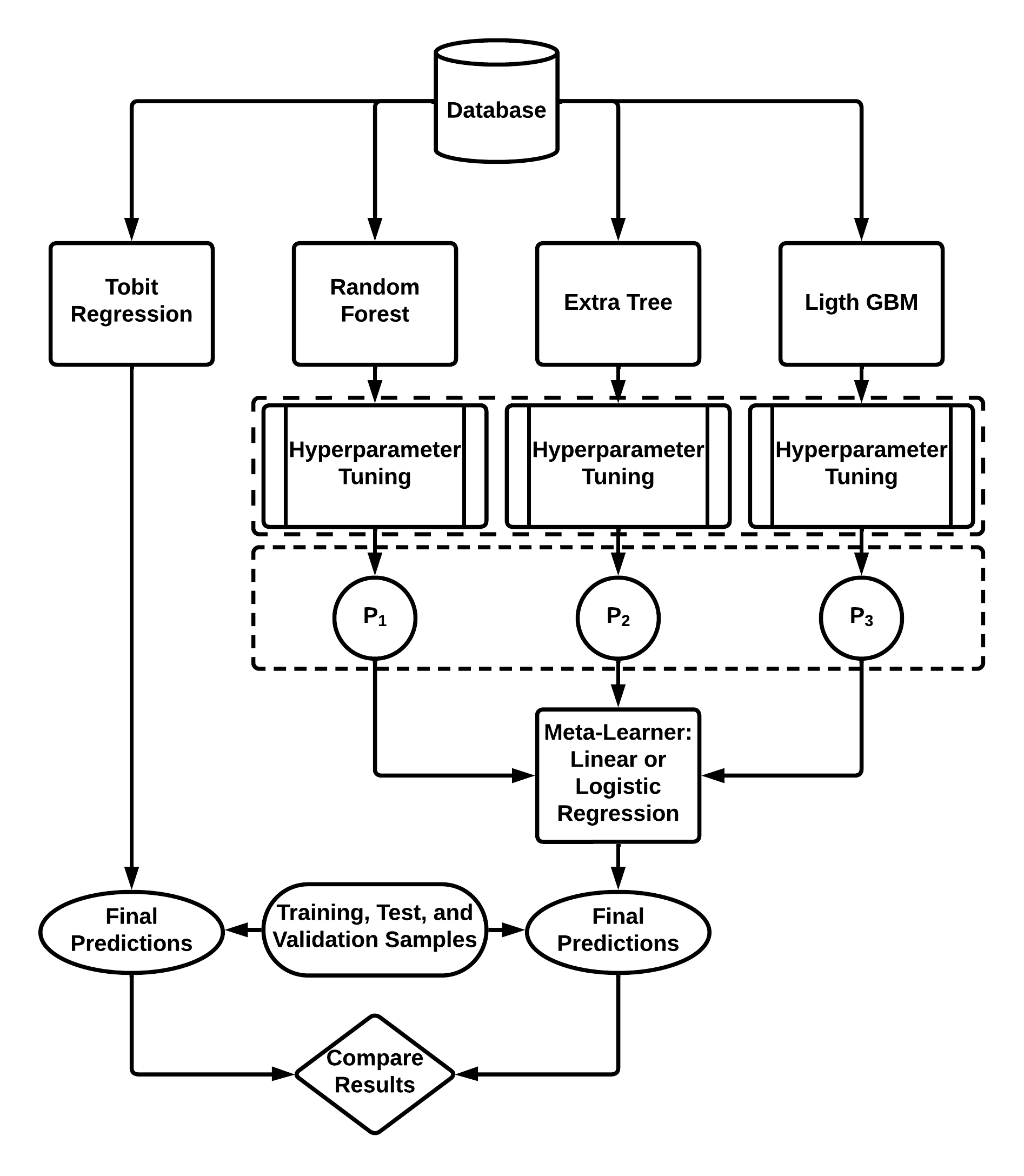}
  \caption{Illustration of the concept of blending}\label{fig:blending}
\end{figure}
With the initial classification and additional attributes from the asset management system, the event then moves to the regression module. The regression module runs a regression model that is fine-tuned to predict the exact duration of the event based on the subset to which the data point belongs. Each model is trained on the training data, tested on holdout or test data, and then retrained on both the train and holdout data before saving the model.

To develop a model with the best efficiency, the solution employs a process called blending, which combines the strengths of multiple models to create a single model with hybrid features. 

In the classification module, it was found that the Random Forest model, the Extra Tree Classifier model, and the Light GBM model achieved the lowest mean absolute error (MAE). The predictions of these three models are combined by a meta-learner to give a blended final prediction, as shown in Figure \ref{fig:blending}. 
Similarly, for the regression module, the predictions from the combinations of Random Forest model in combination with Catboost model are fed into meta-learner for short-duration prediction. Predictions from Random Forest model and Huber model are fed into meta-learner for medium-duration prediction. For long-duration event prediction, XG Boost model alone is used since it performed better than blended models. A Tobit model is used as a reference to evaluate the MAE value and determine how the methodology is improving the results.

\subsection{Model Evaluation Metrics}
One commonly used performance metric for evaluating classification models is the Area Under the Curve (AUC) of the Receiver Operating Characteristics (ROC) plot. The ROC curve plots the true positive rate against the false positive rate at different classification thresholds, and the AUC is calculated by finding the area under the ROC curve. A higher AUC value indicates that the model is better at distinguishing between the classes. In the problem at hand, accurate classification of an event into the correct duration category is crucial because it ensures that the appropriate regression model, which has been specifically trained for that category, is applied.

\[ TPR=\frac{TP}{FP+TN} \] 

\[ FPR=\frac{FP}{FP+TN} \] 


\[ Precision=\frac{TP}{TP+FP} \] 

\[ Recall=\frac{TP}{TP+FN} \]

Additionally, precision, accuracy, and recall are useful parameters for evaluating model performance. While accuracy indicates the performance of the model in predicting the correct class or category, precision indicates the ratio of what the model predicted correctly to what it predicted overall, and recall indicates what the model predicted correctly to what the true classifications are. Along with high AUC values, the best possible value for recall is also desired because the model needs to be sensitive toward the higher severity incidents that are associated with longer durations. In the case of regression models, commonly used evaluation metrics include Mean Absolute Error (MAE) and Root Mean Squared Error (RMSE). 

\[ MAE = (\frac{1}{n})\sum_{i=1}^{n}\left | t_{pi} - t_{oi} \right |\]


\subsection{Model description}
The best-performing machine learning model for the classification and regression modules was selected based on the highest AUC value for the classification module and the lowest mean absolute error (MAE) for the regression module. Tree-based models generally performed better. To further improve model performance, an ensemble technique called blending was applied. The following section provides a general overview of the top-performing models used in the classification and regression modules.

\subsubsection{Random Forest classifier and regressor}
Random Forest is an ensemble method that uses bootstrap aggregation, also known as bagging, to train multiple decision trees on different samples of the data with replacement. The final output is obtained by combining the predictions of these decision trees. Two common hyperparameters used to tune and improve the performance of a Random Forest model are max\_depth and n\_estimators. Max\_depth determines the maximum depth of each tree, while n\_estimators specifies the number of trees in the forest. Adjusting these hyperparameters can help optimize the model's performance.

\subsubsection{CatBoost classifier and regressor}
The CatBoost algorithm is designed to perform well on categorical data and has several advantages. It can produce high-quality results without the need for extensive parameter tuning, it provides fast and improved predictions by reducing overfitting, and it can handle datasets with less data. This model is typically effective on heterogenous data, which is characterized by low quality, high variability of data types, ambiguous data, and data with high redundancy. In our case, the incident duration dataset can be considered heterogenous, and it has been found that the CatBoost model performs well on this type of data.

\subsubsection{Light Gradient Boost classifier and regressor}
Light GBM is a gradient boosting machine learning framework based on tree-based algorithms. It grows the tree leaf-wise, which is different from other tree-based algorithms that grow level-wise. Light GBM has the advantages of high speed and the ability to handle large datasets with lower memory usage and it focuses on the accuracy of the results.

\subsubsection{Blending}
Blending is an ensemble technique that uses a linear regression or logistic regression model as a meta model to blend or combine the predictions of multiple machine learning models to estimate more accurate predictions. The concept of blending has been demonstrated in Figure \ref{fig:blending} where the base ML models RF, ET and Light GBM are individually trained on the training data and make individual predictions on the holdout or test dataset. The meta model, which is usually a logistic regression model, is trained on these predictions to make a single final prediction.

\subsubsection{Huber regressor}
Traditional regression models can produce misleading results if the underlying assumptions are not met. Robust regression methods are designed to minimize the impact of these violations. One commonly used method for robust regression is the Huber loss, which is less sensitive to outliers than the squared error loss. This makes it a useful tool for minimizing the effect of outliers on the regression analysis. The Huber loss function can be defined as below:
\[
    L_{\delta}(y,f(x))= 
\begin{cases}
    \frac{1}{2}{a^2},& \text{for |y-f(x)|} \leq \delta\\
    \delta(|y-f(x)|-\frac{1}{2}\delta),              & \text{otherwise}
\end{cases}
\]

\begin{figure}[t]
    \centering 
\scalebox{0.97}{
\begin{subfigure}{0.56\textwidth}
  \includegraphics[width=\linewidth]{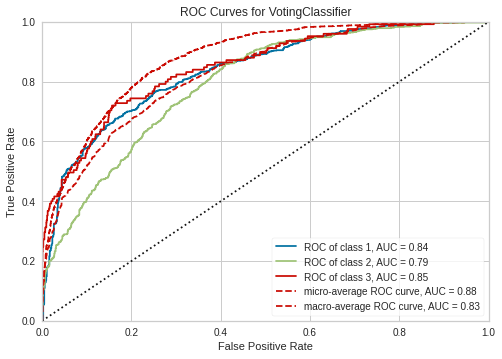}
  \label{fig:AUC_basic}
\end{subfigure}\hfil 
\begin{subfigure}{0.54\textwidth}
  \includegraphics[width=\linewidth]{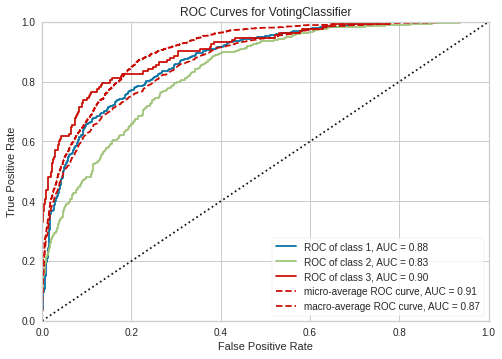}
  \label{fig:AUC_all}
\end{subfigure}}\hfil\label{AUC_all}
\caption{Comparison of AUC (basic feature set(left), all features(right) (validation dataset)}
\label{AUC}
\end{figure}

\subsubsection{Tobit model}
Tobit model is a class of regression models developed in 1958 with the intention of mitigating the problem truncated data. The Tobit model can also be considered a special case of a censored linear regression model that can be represented by the equation below:\\

$$\begin{cases}
    y_{i}^* \quad  if\quad y_{L}<y_{i}^*<y_{U},\\
    y_{L}\quad  if\quad y_{i}^*<y_{L},\\
    y_{U}\quad  if\quad y_{i}^*>=y_{U}\\

\end{cases}$$
Here, $y_{i}^*$ is a latent variable that is not always observable and, $y_{L}$ and $y_{U}$ are the lower and upper limits to which the model is censored.

\section{Results}
During the training phase of the analysis, several models, including Random Forest (RF), CatBoost Regressor, Gradient Boost, Ada Boost, and XGBoost, were trained and tested on the short, medium, and long duration datasets, as well as other base models such as SVM, Decision Tree, and KNN. The models were then used to make predictions on the test dataset and were ranked in descending order of AUC values for classification models and ascending order of MAE values for regression models. The classification model(s) with the highest AUC values and the regression model(s) with the lowest MAE values were identified as the best models. Additionally, blending was performed on the top 2 (BM2), top 3 (BM3), top 4 (BM4), and top 5 (BM5) best machine learning models to create four new models. These models were also evaluated in the same way and compared against the individual models to evaluate the improvement resulting from blending.

This remainder of this section discusses the training, testing and validation results provided in Tables \ref{table:class_results}, \ref{reg_basic_comparison} and \ref{reg_all_comparison}.  

The results were obtained through the process of deciding the ML components of the developed framework under the availability of both the feature sets explained in Table \ref{tab:featuresetinfo}.
The basic feature set includes information collected from the site of the incident by the reporting policeman. This includes information about the time and location of crash, type of crash, vehicle information, involvement of trucks in the incident and the number of injured occupants and so on. This information can be used to make the first prediction. In the next phase, more information relating to the incident is obtained like the AADT, hourly volume, road and environmental factors related to the location of the incident from RAMS, and whether a responder arrived or not at the incident site. This additional information along with the basic features  (full features) helps make a more refined prediction of the time required to clear the incident.
\subsection{Classification Module Results}
\begin{figure}[t]
    \centering 
\scalebox{0.95}{
\begin{subfigure}{0.54\textwidth}
  \includegraphics[width=\linewidth]{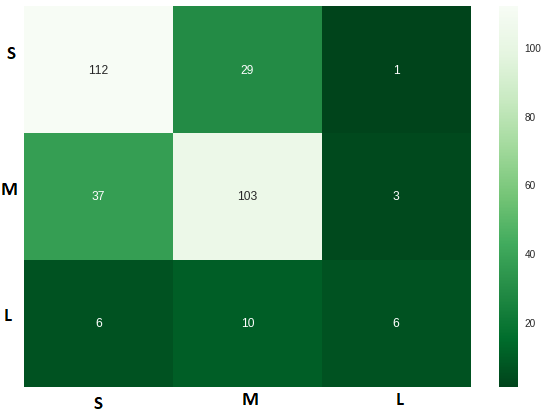}
  \label{confmatrix_basic}
\end{subfigure}\hfil 
\begin{subfigure}{0.54\textwidth}
  \includegraphics[width=\linewidth]{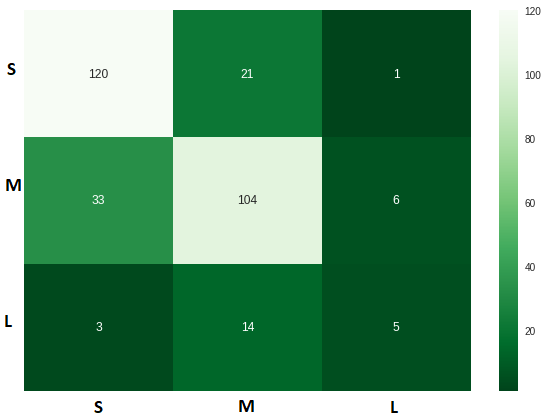}
\end{subfigure}}\hfil 
\caption{Confusion matrices (basic feature set(left), all features(right) (validation dataset) with Short (S), Medium (M), and Long (L) classifications. Observed classifications are on the x-axis and predicted classifications are on the y-axis}
\label{confmatrix}
\end{figure}

In the classification module, a blended model obtained from combining Random Forest model with an Extra Tree model and Light GBM yields the best AUC of 81.9\% for the basic features and an AUC of 86\% when the full features are available as shown in Table \ref{table:class_results}. The corresponding AUC graphs can be found in Figure \ref{AUC}. These models are trained, saved and used to predict on the validation dataset. The results of the prediction are shown in Figure \ref{confmatrix} with S, M and L indicating short, medium and long duration events. It is evident from comparing the results in Table \ref{table:class_results} that the classification accuracy of short duration incidents improved by 7.14\% and medium duration events improved by 0.97\% when additional features are available beyond the basic features.

\begin{table}
    \centering
\caption{Results from classification module (based on FS1 and FS2)}
 \label{table:class_results}
\begin{tabular}{|p{3cm}|p{1.5cm}|p{3cm}|p{1.5cm}|p{1.5cm}|p{1.5cm}|p{1.5cm}|}
\hline
Data&Features& Best Model&AUC &Precision & Accuracy & Recall\\
\hline
Train&Basic&RF+ET+LGBM&0.81 & 0.683 & 0.684 & 0.6026\\
\hline
Test&Basic&RF+ET+LGBM&0.819 & 0.702 & 0.7 & 0.611\\
\hline
Train&All&RF+ET+LGBM&0.87 & 0.76 & 0.753 & 0.662\\
\hline
Test&All&RF+ET+LGBM&0.86 & 0.739 & 0.737 & 0.661\\
\hline
\end{tabular}
\end{table}

\subsection{Regression Module Results}
 
 \begin{table}
    \centering
\caption{Results from regression module (based on basic features)}

\begin{tabular}{|p{2cm}|p{4.5cm}|p{2.5cm}|p{2cm}|p{5cm}|}
\hline
Duration&Model&MAE(test) &MAE(valid) \\
\hline
Short&RF&5.66&24.58 \\
\hline
Medium& RF&14.07&42.27\\
\hline
Long&RF+XGBoost&41.9&129.1\\
\hline
\end{tabular}
 \label{reg_basic_comparison}
\end{table} 

\begin{table}
    \centering
\caption{Results from regression module (based on all features)}

\begin{tabular}{|p{2cm}|p{4.5cm}|p{2.5cm}|p{2cm}|p{4cm}|}
\hline
Duration&Model&MAE(test) &MAE(valid) \\
\hline
Short&RF+Catboost&5.76&22.11 \\
\hline
Medium& RF+Huber&15.73&47.3\\
\hline
Long&XGBoost&33.27&96.09\\
\hline
\end{tabular}
 \label{reg_all_comparison}
\end{table} 
\begin{figure}[h!]
\centering    

\begin{subfigure}{0.45\linewidth}
\begin{tikzpicture}[scale=0.8]
\pgfplotsset{every axis legend/.append style={
		at={(0.5,1.03)},
		anchor=south}}
\begin{axis}[legend columns=4,width=8cm,height=6cm,
xtick=data,
ymax=170
        ,xticklabels={Unsup,Sup\_MC,Tobit\_MC},
        x tick label style={rotate=45, anchor=north east, inner sep=0mm},
        ylabel =MAE,
        ]

    \addplot[mark=diamond*,color=red, dashed] coordinates {(1,13.9)(2,5.66)(3,6.45)};
\addplot[mark=oplus*,color=blue]  coordinates
{(1,22.013)(2,14.07)(3,15.87)};
\addplot[mark=square*,color=black, dotted]  coordinates
{(1,85.471) (2,41.9)(3,82.55)};
\legend{Short,Medium,Long}
\end{axis}
\end{tikzpicture}
\end{subfigure}
\begin{subfigure}{0.45\linewidth}
\begin{tikzpicture}[scale=0.8]
\pgfplotsset{every axis legend/.append style={
		at={(0.5,1.03)},
		anchor=south}}
\begin{axis}[legend columns=4,width=8cm,height=6cm,
xtick=data,
ymax=170
        ,xticklabels={Unsup,Sup\_MC,Tobit\_MC},
        x tick label style={rotate=45, anchor=north east, inner sep=0mm},
        ylabel =MAE,
        ]

\addplot[mark=diamond*,color=red, dashed] coordinates {(1,10.606)(2,5.765)(3,5.99)};
\addplot[mark=oplus*,color=blue]  coordinates
{(1,22.12)(2,15.731)(3,20.67)};
\addplot[mark=square*,color=black, dotted]  coordinates
{(1,86.84) (2,33.273)(3,59.78)};
\legend{Short,Medium,Long}
\end{axis}
\end{tikzpicture}
\end{subfigure}
\caption{Comparison of MAE values (basic feature set(left), all features(right) (test dataset)}
\label{graph_mae_test}
\end{figure}
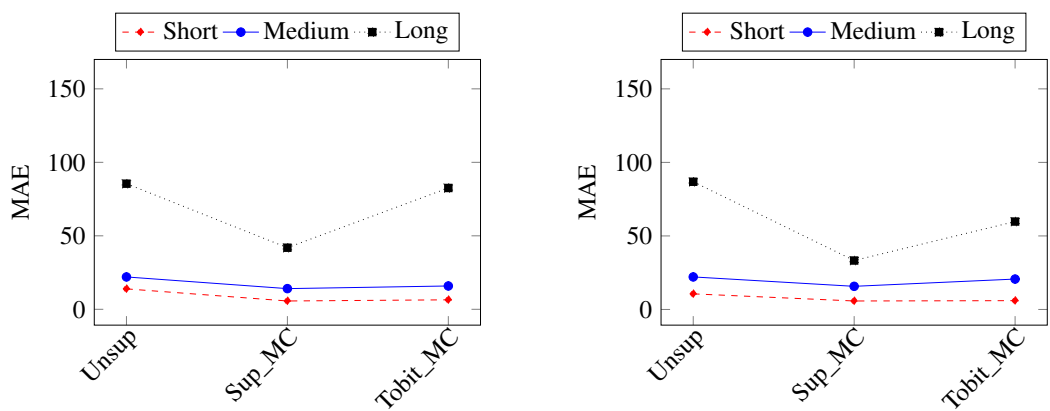

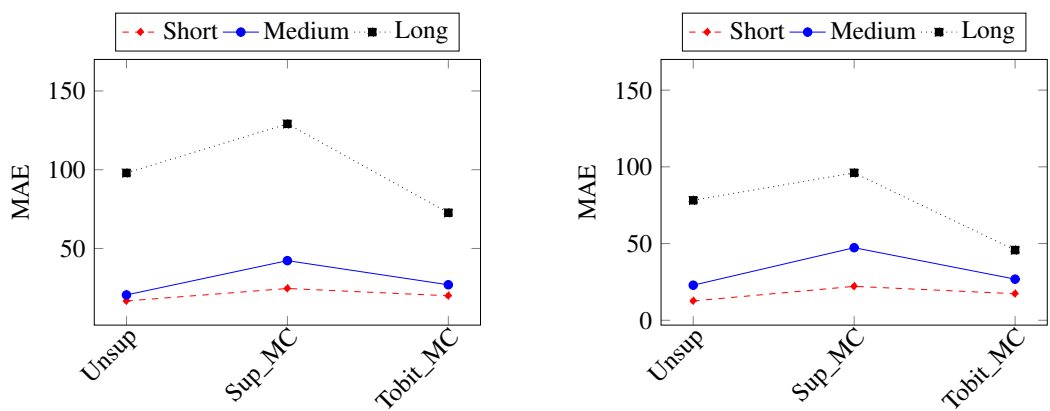
\begin{figure}[h!]
\centering    

\begin{subfigure}{0.45\linewidth}
\begin{tikzpicture}[scale=0.8]
\pgfplotsset{every axis legend/.append style={
		at={(0.5,1.03)},
		anchor=south}}
\begin{axis}[legend columns=4,width=8cm,height=6cm,
xtick=data,
ymax=170
        ,xticklabels={Unsup,Sup\_MC,Tobit\_MC},
        x tick label style={rotate=45, anchor=north east, inner sep=0mm},
        ylabel =MAE,
        ]

    \addplot[mark=diamond*,color=red, dashed] coordinates {(1,16.665)(2,24.583)(3,19.968)};
\addplot[mark=oplus*,color=blue]  coordinates
{(1,20.491)(2,42.27)(3,26.94)};
\addplot[mark=square*,color=black, dotted]  coordinates
{(1,97.946) (2,129.1)(3,72.67)};
\legend{Short,Medium,Long}
\end{axis}
\end{tikzpicture}
\end{subfigure}
\begin{subfigure}{0.45\linewidth}
\begin{tikzpicture}[scale=0.8]
\pgfplotsset{every axis legend/.append style={
		at={(0.5,1.03)},
		anchor=south}}
\begin{axis}[legend columns=4,width=8cm,height=6cm,
xtick=data,
ymax=170
        ,xticklabels={Unsup,Sup\_MC,Tobit\_MC},
        x tick label style={rotate=45, anchor=north east, inner sep=0mm},
        ylabel =MAE,
        ]

\addplot[mark=diamond*,color=red, dashed] coordinates {(1,12.585)(2,22.115)(3,17.3)};
\addplot[mark=oplus*,color=blue]  coordinates
{(1,22.868)(2,47.302)(3,26.81)};
\addplot[mark=square*,color=black, dotted]  coordinates
{(1,78.196) (2,96.094)(3,45.76)};
\legend{Short,Medium,Long}
\end{axis}
\end{tikzpicture}
\end{subfigure}
\caption{Comparison of MAE values (basic feature set(left), all features(right) (validation dataset)}
\label{graph_mae_validation}
\end{figure}
From Table \ref{reg_basic_comparison}, it was found that for short and medium duration incidents, the RF model had the best performance, while a blend model consisting of the RF model and XGBoost performed the best in the case of long duration incidents. The best MAEs reported on the test dataset were 5.66, 14.07, and 41.9 minutes, respectively. When all the variables were considered, the MAE values improved to 5.76, 15.73, and 33.27 minutes, respectively. When the performance of the models on the validation dataset was evaluated, it was found that the MAE for the validation dataset was significantly higher than the MAE of the test dataset. This deviation from the expected values could be due to the large variance in the data. When additional features were available, an improvement in MAE values was observed for long duration incidents, as shown in Table \ref{reg_all_comparison}. In addition to blending, hyperparameter tuning of the best model, as reported by the pycaret package, was also performed using the Optuna library. However, hypertuning did not help improve the performance of the model in any of the cases considered.

\begin{figure}[h!]
 \centering
 \begin{subfigure}[h]{0.49\textwidth}
     \centering
    \begin{tikzpicture}[scale=0.8]
\pgfplotsset{every axis legend/.append style={
		at={(0.5,1.03)},
		anchor=south}}
\begin{axis}[legend columns=4,width=8cm,height=6cm,
xtick=data,
xlabel=Incident Duration,
ymax=180
        ,xticklabels={Short,Medium,Long},
        x tick label style={rotate=0, anchor=north east, inner sep=1mm}, 
        ylabel =MAE (Test) (min),
        ]

       \addplot[mark=diamond*,color=red, dashed] coordinates {(1,13.1)(2,25.1)(3,97.25)};
    \addplot[mark=diamond*,color=blue, dashed] coordinates {(1,5.66)(2,14.071)(3,41.9)};
\legend{Without\_class,With\_class}
\end{axis}
\end{tikzpicture}
     \caption{MAE (test) for developed framework}
     \label{mae_test_dev}
 \end{subfigure}
 \hfill
 \begin{subfigure}[h]{0.49\textwidth}
     \centering
    \begin{tikzpicture}[scale=0.8]
\pgfplotsset{every axis legend/.append style={
		at={(0.5,1.03)},
		anchor=south}}
\begin{axis}[legend columns=4,width=8cm,height=6cm,
xtick=data,
xlabel=Incident Duration,
ymax=180
        ,xticklabels={Short,Medium,Long},
        x tick label style={rotate=0, anchor=north east, inner sep=1mm},
        ylabel =MAE (Validation)(min),
        ]
     \addplot[mark=diamond*,color=red, dashed] coordinates {(1,12.96)(2,23.58)(3,117.24)};
    \addplot[mark=diamond*,color=blue, dashed] coordinates {(1,24.58)(2,42.27)(3,129.1)};
\legend{Without\_class,With\_class}
\end{axis}
\end{tikzpicture}
     \caption{MAE (validation) for developed framework}
     \label{mae_valid_dev}
 \end{subfigure}
 \hfill
 \begin{subfigure}[h]{0.49\textwidth}
     \centering
\begin{tikzpicture}[scale=0.8]
\pgfplotsset{every axis legend/.append style={
		at={(0.5,1.03)},
		anchor=south}}
\begin{axis}[legend columns=4,width=8cm,height=6cm,
xtick=data,
xlabel=Incident Duration,
ymax=180
        ,xticklabels={Short,Medium,Long},
        x tick label style={rotate=0, anchor=north east, inner sep=1mm},
        ylabel =MAE (Test)(min),
        ]

       \addplot[mark=diamond*,color=red, dashed] coordinates {(1,12.43)(2,26.39)(3,163.08)};
    \addplot[mark=diamond*,color=blue, dashed] coordinates {(1,6.45)(2,15.87)(3,82.55)};
\legend{Without\_class,With\_class}
\end{axis}
\end{tikzpicture}
    \caption{MAE (test) for Tobit model}
     \label{mae_test_tobit}
 \end{subfigure}
 \hfill
 \begin{subfigure}[h]{0.49\textwidth}
     \centering
   \begin{tikzpicture}[scale=0.8]
\pgfplotsset{every axis legend/.append style={
		at={(0.5,1.03)},
		anchor=south}}
\begin{axis}[legend columns=4,width=8cm,height=6cm,
xtick=data,
xlabel=Incident Duration,
ymax=180
        ,xticklabels={Short,Medium,Long},
        x tick label style={rotate=0, anchor=north east, inner sep=1mm},
        ylabel =MAE (Validation)(min),
        ]
     \addplot[mark=diamond*,color=red, dashed] coordinates {(1,19.08)(2,27.27)(3,92.97)};
    \addplot[mark=diamond*,color=blue, dashed] coordinates {(1,19.97)(2,26.94)(3,72.67)};
\legend{Without\_class,With\_class}
\end{axis}
\end{tikzpicture}
     \caption{MAE (validation) for Tobit model}
     \label{mae_valid_tobit}
 \end{subfigure}
   \caption{Comparison of MAE values corresponding to test and validation datasets for framework model and Tobit model with and without prior classification when basic feature set available}
    \label{testandvalid_basic}
\end{figure}
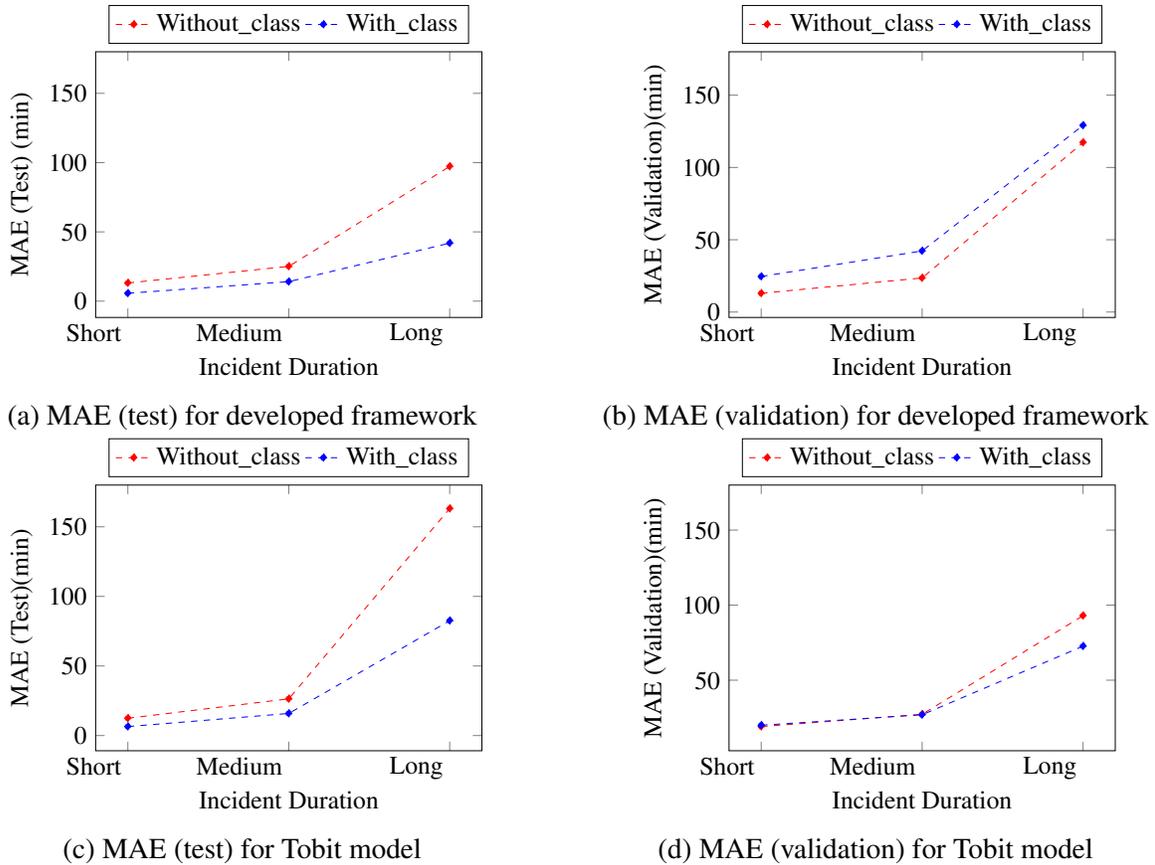

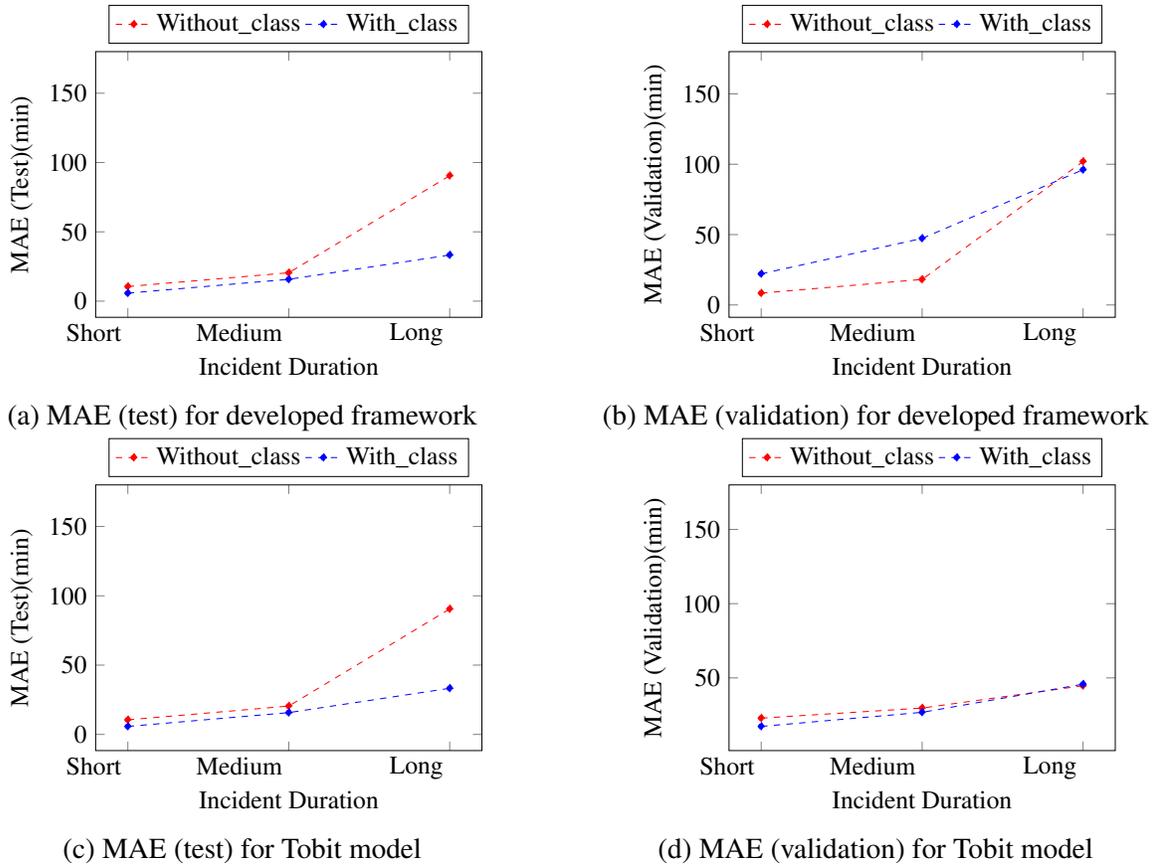
\begin{figure}[h!]
 \centering
 \begin{subfigure}[h]{0.49\textwidth}
     \centering
   \begin{tikzpicture}[scale=0.8]
\pgfplotsset{every axis legend/.append style={
		at={(0.5,1.03)},anchor=south}}
\begin{axis}[legend columns=4,width=8cm,height=6cm,
xtick=data,
xlabel= Incident Duration,
ymax=180
        ,xticklabels={Short,Medium,Long},
        x tick label style={rotate=0, anchor=north east, inner sep=1mm},
        ylabel =MAE (Test)(min),
        ]
       \addplot[mark=diamond*,color=red, dashed] coordinates {(1,10.52)(2,20.46)(3,90.50)};
    \addplot[mark=diamond*,color=blue, dashed] coordinates {(1,5.76)(2,15.73)(3,33.27)};
\legend{Without\_class,With\_class}
\end{axis}
\end{tikzpicture}
     \caption{MAE (test) for developed framework}
     \label{mae_test_dev_full}
 \end{subfigure}
 \hfill
 \begin{subfigure}[h]{0.49\textwidth}
     \centering
   \begin{tikzpicture}[scale=0.8]
\pgfplotsset{every axis legend/.append style={
		at={(0.5,1.03)},
		anchor=south}}
\begin{axis}[legend columns=4,width=8cm,height=6cm,
xtick=data,
xlabel= Incident Duration,,
ymax=180
        ,xticklabels={Short,Medium,Long},
        x tick label style={rotate=0, anchor=north east, inner sep=1mm},
        ylabel =MAE (Validation)(min),
        ]
     \addplot[mark=diamond*,color=red, dashed] coordinates {(1,8.399)(2,18.144)(3,101.951)};
    \addplot[mark=diamond*,color=blue, dashed] coordinates {(1,22.115)(2,47.302)(3,96.093)};
\legend{Without\_class,With\_class}
\end{axis}
\end{tikzpicture}
     \caption{MAE (validation) for developed framework}
     \label{mae_valid_dev_full}
 \end{subfigure}
 \hfill
 \begin{subfigure}[h]{0.49\textwidth}
     \centering
\begin{tikzpicture}[scale=0.8]
\pgfplotsset{every axis legend/.append style={
		at={(0.5,1.03)},
		anchor=south}}
\begin{axis}[legend columns=4,width=8cm,height=6cm,
xtick=data,
xlabel= Incident Duration,
ymax=180
        ,xticklabels={Short,Medium,Long},
        x tick label style={rotate=0, anchor=north east, inner sep=1mm},
        ylabel =MAE (Test)(min),
        ]

       \addplot[mark=diamond*,color=red, dashed] coordinates {(1,10.523)(2,20.46)(3,90.503)};
    \addplot[mark=diamond*,color=blue, dashed] coordinates {(1,5.765)(2,15.731)(3,33.273)};
\legend{Without\_class,With\_class}
\end{axis}
\end{tikzpicture}
    \caption{MAE (test) for Tobit model}
     \label{mae_test_tobit_full}
 \end{subfigure}
 \hfill
 \begin{subfigure}[h]{0.49\textwidth}
     \centering
  \begin{tikzpicture}[scale=0.8]
\pgfplotsset{every axis legend/.append style={
		at={(0.5,1.03)},
		anchor=south}}
\begin{axis}[legend columns=4,width=8cm,height=6cm,
xtick=data,
xlabel= Incident Duration,
ymax=180
        ,xticklabels={Short,Medium,Long},
        x tick label style={rotate=0, anchor=north east, inner sep=1mm},
        ylabel =MAE (Validation)(min),
        ]
     \addplot[mark=diamond*,color=red, dashed] coordinates {(1,22.88)(2,29.686)(3,44.713)};
    \addplot[mark=diamond*,color=blue, dashed] coordinates {(1,17.3)(2,26.81)(3,45.76)};
\legend{Without\_class,With\_class}
\end{axis}
\end{tikzpicture}
     \caption{MAE (validation) for Tobit model}
     \label{mae_valid_tobit_full}
 \end{subfigure}
   \caption{Comparison of MAE values corresponding to test and validation datasets for framework model and Tobit model with and without prior classification when full feature set is available}
    \label{testandvalid_all}
\end{figure}

\subsection{Evaluation of Classification Module in Developed Framework}
To understand the usefulness of the classification module in the developed framework, a Random Forest model was used to predict incident duration without performing any prior classification (indicated as Without\_class). The mean absolute error (MAE) values obtained from the test, and validation datasets were compared in Figure \ref{testandvalid_basic} and Figure \ref{testandvalid_all}. It was observed that for the basic feature set, including the classification step prior to regression helped to reduce the MAE of the model for short, medium, and long-duration incidents by 56\%, 39.76\%, and 57.131\%, respectively, as evident in the test data results shown in Figure \ref{mae_test_dev}. When all the features were considered, prior classification helped to reduce MAE by 45.2, 23.1, and 63.2\% for the respective duration categories, as shown in Figure \ref{mae_test_dev_full}. However, when considering the validation dataset, the overall MAE was found to improve only for the long-duration dataset when all the features were considered together, as shown in Figures \ref{mae_valid_dev} and \ref{mae_valid_dev_full}.

The same analysis was applied to the Tobit model, and it was found that the pre-classification model improved the MAE values for both the basic and all feature sets in both the test and validation datasets, as shown in Figures \ref{mae_test_tobit}, \ref{mae_valid_tobit}, \ref{mae_test_tobit_full}, \ref{mae_valid_tobit_full} with the exception of two cases.


\subsection{Comparison of our framework with unsupervised approach and Tobit model}\label{sec:comparison}

For the unsupervised approach, K-means clustering model was applied instead of supervised classification. The optimal number of clusters for both the basic feature set and the feature set involving all the features was found to be 4. The optimal number of clusters was chosen based on the elbow fit method and the Silhouette Factor. The Silhouette coefficient obtained for both the cases have been presented in Table \ref{table:silhoette}. A higher value closer to 1 obtained when more features are available gives better distinction between clusters. 
\begin{table}
    \centering
\caption{Comparison of Silhouette factors and optimal number of clusters}
 \label{table:silhoette}
\begin{tabular}{|p{4cm}|p{4cm}|p{4cm}|}
\hline
Feature Set & Number of clusters & Silhouette factor\\
\hline
Basic &4&0.07 \\
\hline
All &4&0.81\\
\hline
\end{tabular}
\end{table}

The MAE values obtained from the supervised and unsupervised approaches for the test and validation datasets have been plotted in Figures \ref{graph_mae_test} and \ref{graph_mae_validation}. "Unsup", "Sup\_MC", and "Tobit\_MC" stand for the MAE in the unsupervised approach, the supervised approach with misclassification error, and the Tobit model with misclassification error, respectively. In all three cases for the test dataset, the MAE values obtained for our framework are lower than those for the unsupervised and Tobit approaches. As the number of available features increases, the classification error decreases, resulting in a lower MAE value. This effect is more pronounced in the case of long duration incidents. However, for the validation dataset, our framework performs slightly worse than the unsupervised approach and Tobit model. This is more noticeable in the case of medium duration incidents (30 minutes to 2 hours).

\section{Discussion}
In this section, the results of the classification and regression models that were developed and tested in the previous section are discussed. The motivation for choosing certain models or combinations of models over others is also explained.

A variety of models were tested for the classification module and ranked based on the highest AUC for both the basic and full feature sets. Ultimately, the model that combines a random forest, an extra tree, and a light gradient boosting machine had the highest AUC and was selected for the classification module. For the regression module, models were selected based on the lowest MAE, which varied based on the initiation duration classification from the classification module and the basic or full feature set. When only basic features are available, Table \ref{reg_basic_comparison} shows the selected models with the lowest MAE for the short, medium, and long classifications. When the full feature set is available, the best regression models for the short, medium, and long durations are shown in Table \ref{reg_all_comparison}. The combination of the classification and regression modules combine the best models for estimating the incident duration based on the pre-trained dataset.

For validating the results against other models, the performance of the developed supervised framework was compared to two additional frameworks that included an unsupervised clustering with regression and a Tobit model. The mean absolute error (MAE) was used as the evaluation metric for comparison. The results of the test data showed that the supervised framework performs better than the unsupervised approach and the Tobit model. With the validation dataset, the performance of the supervised framework did not match the improved performance seen on the test dataset. However, the Tobit model with the pre-classification module showed improved efficiency on both the test and validation datasets. The Tobit model's performance was similar to that of the supervised framework only when the pre-classification module was present.

It was also observed that for short and medium-duration data, all three approaches produced similar MAE values. However, the pre-classification module improved the results of long-duration events more than the other two categories.

Overall, the Tobit model's results were similar to those of the supervised network's regression module, and it produced improved results on both the test and validation datasets. Therefore, there may be value in considering a simple Tobit model after the classification module.

The performance of the models in the classification and regression modules was evaluated using the area under the curve (AUC) and mean absolute error (MAE), respectively. The highest AUC achieved was 86\% when additional variables, such as road characteristics and responder information, were included. In terms of MAE, the lowest values achieved were 5.66 minutes for short-duration events, 14.07 minutes for medium-duration events, and an average of 41.9 minutes for long-duration events on the test dataset when using only basic features. When additional features were available, the MAE increased to 5.76 and 15.73 for short and medium incidents, respectively, but decreased to 33.27 minutes for long-duration incidents.

To compare the results to those obtained in previous research and to understand if the findings were consistent with those reported in the literature, the mean absolute percentage error (MAPE) values were also calculated for each of the models. The MAPE values achieved with the basic dataset were 99.16\% for short duration, 24.45\% for medium duration, and 18.34\% for long duration. The inclusion of additional variables, such as road characteristics and highway helper information, resulted in improved MAPE values of 96.88\% and 16.21\% for short and long incidents, respectively.

In a study involving the dynamic prediction of incident duration using an adaptive feature set, the MAPE values reported for short-duration events were 100.9\% for incidents lasting 5-15 minutes and between 75\% and 96\% for incidents lasting 16-35 minutes \cite{9}. For medium-duration events lasting 36-200 minutes, a MAPE between 20\% and 50\% was reported. In comparison, the methods used in this paper achieved MAPE values in the range of 16\% to 26\% for events of a wide range of durations lasting from 30 minutes up to a day when all variables were considered, including variables that are likely not available at the initial stages of responding to incidents (included to ensure comparability with previous research). The MAPE, MAE, and mean absolute relative error (MARE) results reported by other researchers working in this area have been provided in Table \ref{tab:ourresults} for comparison.

\noindent\begin{table}[t]
\centering
\caption{Comparison of our results with existing studies }
 
\scalebox{0.7}{
\begin{tabular}{|p{8cm}|p{14cm}|}
 \hline
 Reference  & Results\\
 \hline
Khattak et al. \cite{21} & MAPE 5-15min:329\%, >120min:80\%
\\
\hline
Valenti et al. \cite{22} & MAPE ANN:44\%,SVR:36\%
\\
\hline
Perera et al. \cite{1} & MAPE between 100\% and 40\%
\\
\hline
Qing et al. \cite{23} & KNN : 59.2\%,CART: 57.1\% Quantile Reg :49.1\%
\\ \hline
Tang et al. \cite{12} & MAPE : 34.8\%
\\ \hline
Hamad et al. \cite{13} & MAE : 36.52 min
\\ \hline
Park et al. \cite{14} & MAE : 0.18 to 0.29
\\ \hline
Ma et al. \cite{15} & MARE : 16.44\%(<15 min) , 33.13\%(>15 min)
\\ \hline
Banishree  et al. \cite{15} & MAPE : 100.9\%(5-15 min) , 75-96\%(16-35 min),20-50\%(>200 min),61\% o 27.58\%(overall)\\
\hline \hline
\textbf{Our Performance (basic features), MAPE} & Short:99.16\%(< 30min), Medium:24.45\%(30-120 min), Long:18.34\%(>120 min)\\
\hline
\textbf{Our Performance (all features), MAPE}&Short:96.88\% ,Medium:26.4\% ,Long:16.21\%
\\
\hline
\textbf{Our Performance (basic features), MAE} & Short:5.66 min , Medium: 14.07 min, Long: 41.9 min\\
\hline
\textbf{Our Performance (all features), MAE}& Short: 5.76 min , Medium: 15.73 min , Long: 33.27 min\\
\hline
\end{tabular}}
\label{tab:ourresults}   
\end{table}

\section{Conclusions and Limitations}
In this research, an ML-based classification and regression framework was proposed for incident duration prediction. The developed framework is capable of generating real-time predictions of incident duration based on the information provided to the response agency in the initial incident call. It additionally integrates other factors including road type, surface type, AADT, hourly volume information, etc., that are not available directly from the incident report enhancing the accuracy of prediction of the incident duration. The predicted incident duration can be used for developing a simple priority-based ranking system that helps traffic operators know which incidents to prioritize and what resources will likely be required. The incident duration information can also be combined with other information, such as the Level of Service (LOS) information, to generate additional actionable insights. 

The framework utilized in this research involved two modules: a classification followed by regression; the classification to provide a quick estimation of the incident duration using a supervised approach and the regression to provide a more accurate prediction as shown in Figure \ref{fig:overallworkflow}.

The success of the incident duration prediction has a strong dependence on the performance of the machine learning models. Among the various ML models trained, tested, and validated, the Random Forest model exhibited high efficiency and consistent performance across all the case scenarios tested. The ensembling technique of blending models exhibiting the lowest MAE values further improved the predictions. 

Overall, the results indicate that the prediction of incident duration is more accurate when there is a pre-classification module that helps determine the class of incident duration. Though an unsupervised classification of events is an option to consider, it is not performing consistently across the three classes. The authors feel that the predictions benefit from the presence of the supervised classification module. In the regression module, however, the Tobit model performs almost on par with other single and blended regression models, as discussed earlier. 

One of the challenges faced in developing and implementing the analytical framework was the skewness of the incident duration data. This was addressed by implementing a box-cox transformation on the target variable. The range of values of incident duration was large: therefore, applying a single model for the whole dataset was not effective. This challenge was addressed by including a classification module to assign incidents to 3 ranges - short (<30 minutes), medium (30 minutes - 2 hours), and long(>2 hours). Categorical factors were one hot encoded, and textual data (e.g., responder information) was extracted and converted to categorical variables for predictive model development. 

A limitation of this research is the use of data for a single state (Iowa). While the results are validated for Iowa, other states may have different trainings, and TIM resources and plans - leading to the results obtained in this research potentially not being applicable to other locations. Thus, future research should apply the framework and methods used in this paper to additional datasets from other geographical areas. Additionally, future research could evaluate the temporal transferability and stability of prediction models developed using this framework and methods. 

\newpage

\bibliographystyle{trb}
\bibliography{trb_template}
\end{document}